\documentclass[final,1p,times]{elsarticle}

\usepackage{amssymb}
\usepackage{amsmath}
\usepackage{amsthm}
\usepackage{xcolor}
\usepackage{rotating}
\usepackage[table]{xcolor}
\usepackage{graphicx}
\usepackage{dsfont}

\begin{document}
\begin{frontmatter}

\title{Characterization and forecasting of national-scale solar power ramp events}

\author[label1]{Luca Lanzilao}
\author[label1,label2]{Angela Meyer}
\affiliation[label1]{organization={School of Engineering and Computer Science, Bern University of Applied Sciences},
             addressline={Quellgasse 21},
             city={Biel},
             postcode={2501},
             state={Bern},
             country={Switzerland}}
\affiliation[label2]{organization={Department of Geosciences and Remote Sensing, TU Delft},
             addressline={Stevinweg 1},
             city={Delft},
             postcode={2628 CN},
             state={South-Holland},
             country={The Netherlands}}

\begin{abstract}
The rapid growth of solar energy is reshaping power system operations and increasing the complexity of grid management. As photovoltaic (PV) capacity expands, short-term fluctuations in PV generation introduce substantial operational uncertainty. At the same time, solar power ramp events intensify risks of grid instability and unplanned outages due to sudden large power fluctuations. Accurate identification, forecasting and mitigation of solar ramp events are therefore critical to maintaining grid stability. In this study, we analyze two years of PV power production from 6434 PV stations at 15-minute resolution. We develop quantitative metrics to define solar ramp events and systematically characterize their occurrence, frequency, and magnitude at a national scale. Furthermore, we examine the meteorological drivers of ramp events, highlighting the role of mesoscale cloud systems. In particular, we observe that ramp-up events are typically associated with cloud dissipation during the morning, while ramp-down events commonly occur when cloud cover increases in the afternoon. Additionally, we adopt a recently developed spatiotemporal forecasting framework to evaluate both deterministic and probabilistic PV power forecasts derived from deep learning and physics-based models, including SolarSTEPS, SHADECast, IrradianceNet, and IFS-ENS. The results show that SHADECast is the most reliable model, achieving a CRPS 10.8\% lower than that of SolarSTEPS at a two-hour lead time. Nonetheless, state-of-the-art nowcasting models struggle to capture ramp dynamics, with forecast RMSE increasing by up to 50\% compared to normal operating conditions. Overall, these results emphasize the need for improved high-resolution spatiotemporal modelling to enhance ramp prediction skill and support the reliable integration of large-scale solar generation into power systems.
\end{abstract}

\end{frontmatter}

\section{Introduction}\label{sec:intro}
Rapid fluctuations in surface solar irradiance (SSI) at local to regional scales can lead to pronounced variability in photovoltaic (PV) power output. This variability, largely governed by the dynamic evolution and motion of cloud systems, represents a major challenge for power system operation and the continued expansion of solar energy \cite{Morales2013, IEA2024}. The problem is especially pronounced for large utility-scale PV plants and regions with a high spatial density of distributed installations, where irradiance variations can translate into significant aggregate power ramps \cite{Tuohy2015}. As solar penetration continues to grow, such ramp events are expected to occur more often and with greater magnitude \cite{Goncalves2024}. These rapid transitions increase the complexity of maintaining grid stability, scheduling reserves, and operating in electricity markets. Consequently, reliable detection and short-term prediction of solar ramp events are crucial for improving operational flexibility, minimizing balancing requirements, and enabling a more efficient coordination between generation assets, grid infrastructure, and electricity demand \cite{Abuella2019, Jie2025}. 

Prior work on solar ramp events has primarily focused on identifying their atmospheric drivers and developing operational countermeasures \cite{Florita2013}. These approaches, including model predictive control \cite{Samu2021} and robust optimization strategies \cite{Pan2024}, are essential for mitigating rapid power fluctuations caused by ramp events that may threaten grid stability \cite{Alam2014, Shivashankar2016, Shivashankar2018}. Observational studies employing all-sky cameras \cite{Haoran2021, Paletta2023, Ruan2024}, Doppler radar measurements, and satellite imagery have investigated cloud processes responsible for abrupt irradiance fluctuations \cite{Si2021}. Anticipating such rapid variability through intra-hour forecasting has therefore become a central research objective \cite{Jie2025, Abuella2018, Zhu2019}. Numerous approaches have been proposed, ranging from numerical weather prediction (NWP) systems to satellite-based retrievals, ground-based sensing networks, and purely data-driven or hybrid modelling frameworks \cite{Yang2022}. The effectiveness of these methods strongly depends on the spatial and temporal scales of interest, with recent advances in deep learning–based spatiotemporal architectures yielding notable improvements in short-term irradiance prediction, particularly when applied to satellite-derived fields \cite{Carpentieri2025, Lanzilao2026}. 

Despite these advances, many existing studies rely on geographically limited case studies or relatively small fleets of PV installations. For instance, \cite{Nouri2024} analyzed data from eight meteorological stations in Spain to introduce a novel ramp-rate metric designed to evaluate ramp events between successive forecast lead times. Similarly, \cite{Florita2013} proposed an automated ramp detection framework based on a single-parameter swinging door algorithm and introduced a dedicated assessment metric using observations from two solar stations in Hawaii. More recently, optical flow techniques and machine learning approaches have been increasingly adopted for forecasting ramp-up and ramp-down events. \cite{Hendrikx2024} employed an all-sky imager in combination with in situ sensor measurements and external meteorological data. They extracted image-derived features such as cloud pixel fraction, brightness, edges, and corner points, and applied optical flow methods to estimate cloud motion and velocity. In a related study, \cite{Wen2021} developed a multi-step forecasting framework for PV ramp-rate control, utilizing a ResNet-18 model for time-series irradiance prediction and incorporating stacked sky images to better capture cloud dynamics and improve short-term forecast accuracy. Likewise, \cite{Chu2015} implemented a low-cost sky-imaging camera network coupled with neural networks to predict very short-term solar irradiance ramp events at two locations in California, USA. 

Despite the rapidly increasing penetration of PV systems and their growing importance for distribution and transmission network operations, solar ramp events have received comparatively less attention than wind ramp events, particularly at large spatial scales \cite{Jie2025, Gallego2015}. To address this gap, we analyze two years of operational PV power measurements at 15-minute resolution from 6434 PV systems distributed across Switzerland, representing, to the best of our knowledge, the first national-scale study of solar ramp events based on PV generation data. We introduce a systematic methodology to detect ramp events, quantify their frequency and occurrence. Moreover, we use satellite-derived SSI fields to characterize the meteorological conditions that promote the development of ramp events. Furthermore, we adopt the spatiotemporal PV forecasting framework developed by \cite{Lanzilao2026} to benchmark multiple forecasting models. The comparison is explicitly conducted in the presence of ramp and non-ramp events, enabling us to assess not only overall forecast skill but also reliability, sharpness, and degradation of performance during highly variable weather conditions. The evaluated satellite-based models include the probabilistic optical-flow model SolarSTEPS \cite{Carpentieri2023}, the probabilistic deep generative diffusion model SHADECast \cite{Carpentieri2025}, and the deterministic deep learning model IrradianceNet \cite{Nielsen2021}. Additionally, we also include the high-resolution integrated forecast system (IFS-ENS) model developed by the European Centre for Medium-Range Weather Forecasts (ECMWF), widely used for SSI forecasting in operational practice \cite{Yang2022b, Wang2022, Sebastianelli2024}.

The main contributions of this study are twofold: (1) we provide a comprehensive national-scale characterization of solar ramp events across more than 6400 PV systems together with a visualization of mesoscale cloud impacts on countrywide PV power production, and (2) we present the first systematic assessment of how different satellite-based and NWP models perform during ramp and non-ramp conditions. The remainder of this article is structured as follows. Section \ref{sec:data} describes the satellite-derived SSI dataset, the PV power dataset, and the forecasting models. Section \ref{sec:ramp_events} introduces the ramp detection methodology and analyses their frequency, spatial characteristics, and meteorological drivers. Section \ref{sec:methodology} details the intraday forecasting framework for both SSI and PV power. Section \ref{sec:results} evaluates the forecast skill of the benchmarked models, discussing their performance under both ramp and non-ramp conditions. Finally, Section \ref{sec:conclusions} summarizes the main findings and outlines future research directions.

\section{Datasets and models}\label{sec:data}
\begin{figure}[t]
	\centering
	\includegraphics[width=1\textwidth]{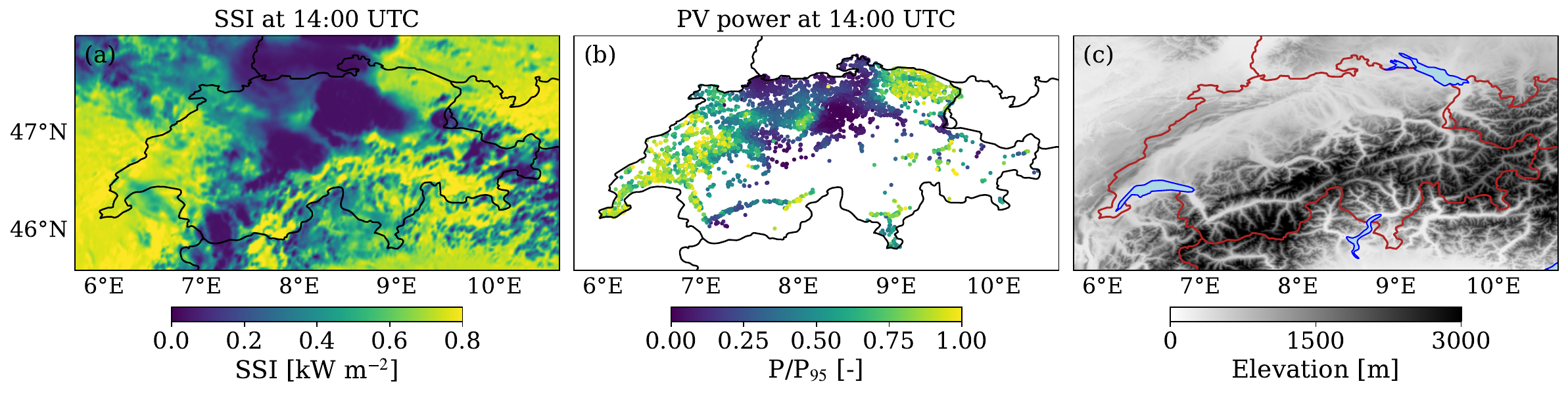}
	\caption{ (a) Satellite-based SSI field and (b) PV power output of the 6434 PV stations, observed on 26 June 2020 at 14:00 UTC. Note that the PV power output is normalized using the station-specific 95th percentile of the power time series. (c) Elevation map showing values in meters above sea level. The black and red lines denote national borders.}
	\label{fig:dataset}
\end{figure}

Our analysis relies on satellite-derived SSI observations and operational PV power measurements. The SSI data are obtained from the Spinning Enhanced Visible and Infrared Imager (SEVIRI) onboard the Meteosat Second Generation (MSG) satellite, positioned in geostationary orbit at 0° longitude \cite{Schmetz2002}. SEVIRI provides full-disk Earth scans with a nadir spatial sampling distance of approximately 3 km. The irradiance fields are retrieved from the EUMETSAT Climate Monitoring Satellite Application Facility (CM SAF), specifically from the High-Resolution European Surface Solar Radiation Data Record (HANNA) \cite{Hanna2025}. The dataset is projected onto a regular grid with a spatial resolution of 0.01° in both longitude and latitude, covering the domain [15°W, 35°E] × [30°N, 70°N]. The SSI fields are available for 2019–2020 at a temporal resolution of 15 minutes \cite{Hanna2025}. Although the temporal coverage of the HANNA product is limited, we chose HANNA over the Surface Radiation Dataset Heliosat (SARAH-3) because the latter underestimates SSI values in snow-covered regions, such as the Alpine arc \cite{Carpentieri2023, Pfeifroth2024}. Figure~\ref{fig:dataset}(a) displays the HANNA SSI field observed on 26 June 2020 at 14:00 UTC over the domain of interest. The high spatial resolution allows for a sharp representation of cloud structures, identifiable by their radiative effect on the surface as areas with lower SSI values. 

The PV data consist of 15-minute production time series from 6434 operational PV systems distributed across Switzerland. The proprietary production data were obtained under a dedicated data usage agreement. A detailed description of the quality control and filtering procedures applied to the raw measurements is provided in \cite{Lanzilao2026}. Assuming an average of 12 daylight hours per day and considering the two-year analysis period (i.e. 2019-2020), the dataset comprises approximately 225 million PV production data points. To enable consistent comparison across installations with different nominal capacities, PV power at each station is normalized with $P_{95}$, defined as the station-specific 95th percentile of its power time series. This normalization ensures comparability across systems and reduces sensitivity to rare extreme outliers. Although normalization by nominal capacity would be a natural alternative, this information is not available. The majority of stations (approximately 76\%) exhibit $P_{95}$ values between 10 and 200 kW, while only 4\% exceed 300 kW, indicating that the dataset is dominated by small- to medium-scale installations. Figure~\ref{fig:dataset}(b) illustrates the locations and corresponding power output of the 6434 PV stations included in this study, recorded on 26 June 2020 at 14:00 UTC while Figure~\ref{fig:dataset}(c) shows the elevation map over the domain of interest. Most PV installations are located in the Swiss Plateau or in valleys within the Alpine region. The cloud patterns visible in Figure~\ref{fig:dataset}(a) are mirrored in the PV power measurements. In fact, regions with reduced PV power output correspond closely to areas with low SSI, highlighting the strong relationship between solar irradiance and power generation. This demonstrates that geostationary satellite observations can effectively capture the spatial footprint of cloud cover and its immediate impact on country-wide solar energy production, providing both high spatial accuracy and low-latency information.

The forecasting models evaluated in this study include the probabilistic optical-flow model SolarSTEPS \cite{Carpentieri2023}, the probabilistic deep generative diffusion model SHADECast \cite{Carpentieri2025}, the deterministic deep learning model IrradianceNet \cite{Nielsen2021}, and the physics-based IFS-ENS model \cite{Roberts2018}. SolarSTEPS is a probabilistic nowcasting framework for satellite-derived clear-sky index (CSI) fields introduced by \cite{Carpentieri2023}. The model first estimates cloud motion vectors (CMVs) using the Lucas–Kanade optical flow method. CSI fields are then decomposed into spatial scales via Fast Fourier Transform, and each scale is forecast using a separate autoregressive (AR) model with spatially correlated noise to generate ensemble members. The resulting components are subsequently recombined and advected to produce the final probabilistic output. In this work, we adopt the same architecture and pre-trained weights as described in \cite{Carpentieri2023}. We additionally evaluate a simplified variant, SolarSTEPS-pa, representing a pure advection (PA) approach. Unlike the full SolarSTEPS framework, this version does not model temporal variability in CSI. Instead, it perturbs the estimated CMVs and advects the CSI fields accordingly. Consequently, SolarSTEPS-pa isolates the contribution of cloud advection while omitting scale decomposition and AR-based temporal modelling. IrradianceNet is a deterministic spatiotemporal deep-learning model based on Convolutional Long Short-Term Memory (ConvLSTM) layers, developed by \cite{Nielsen2021}. It forecasts satellite-derived CSI fields using only past CSI inputs. Originally trained on SARAH-2.1 and subsequently retrained on HelioMont data \cite{Stockli2013, Castelli2014} with an extended autoregressive forecast horizon, it serves as a strong deterministic baseline. However, it does not provide probabilistic forecasts or uncertainty quantification with reasonable computational effort. We adopt the same architecture and pre-trained weights as described in \cite{Carpentieri2025}. SHADECast is a deep generative diffusion model for probabilistic intraday solar forecasting \cite{Carpentieri2025}. The framework first compresses CSI sequences using a variational autoencoder, predicts their deterministic evolution through a latent nowcaster based on Adaptive Fourier Neural Operator blocks and a temporal transformer, and subsequently applies a conditional latent diffusion model to generate ensemble forecasts. The latent outputs are finally decoded into physical CSI fields. We use the architecture and pre-trained weights described in \cite{Carpentieri2025}.

To benchmark these satellite-based methods against a physics-based approach, we include probabilistic SSI forecasts from the physics-based IFS-ENS model \cite{Roberts2018}. Ten randomly selected ensemble members at approximately 9 km spatial resolution are used to ensure consistency with the ensemble size of the satellite-based probabilistic models. Additionally, we apply a machine learning–based bias correction using a U-Net architecture, referring to the corrected forecasts as IFS-ENS-corrected. Further details on the correction methodology and training procedure are provided in \cite{Lanzilao2026}. Overall, seven distinct methods are considered, covering both deterministic and probabilistic forecasting approaches. It is important to emphasize that we do not develop specialized ramp-event forecasting models. Instead, our objective is to assess how state-of-the-art generic spatiotemporal SSI forecasting models perform in capturing solar ramp events.

\begin{figure}[t]
	\centering
	\includegraphics[width=1\textwidth]{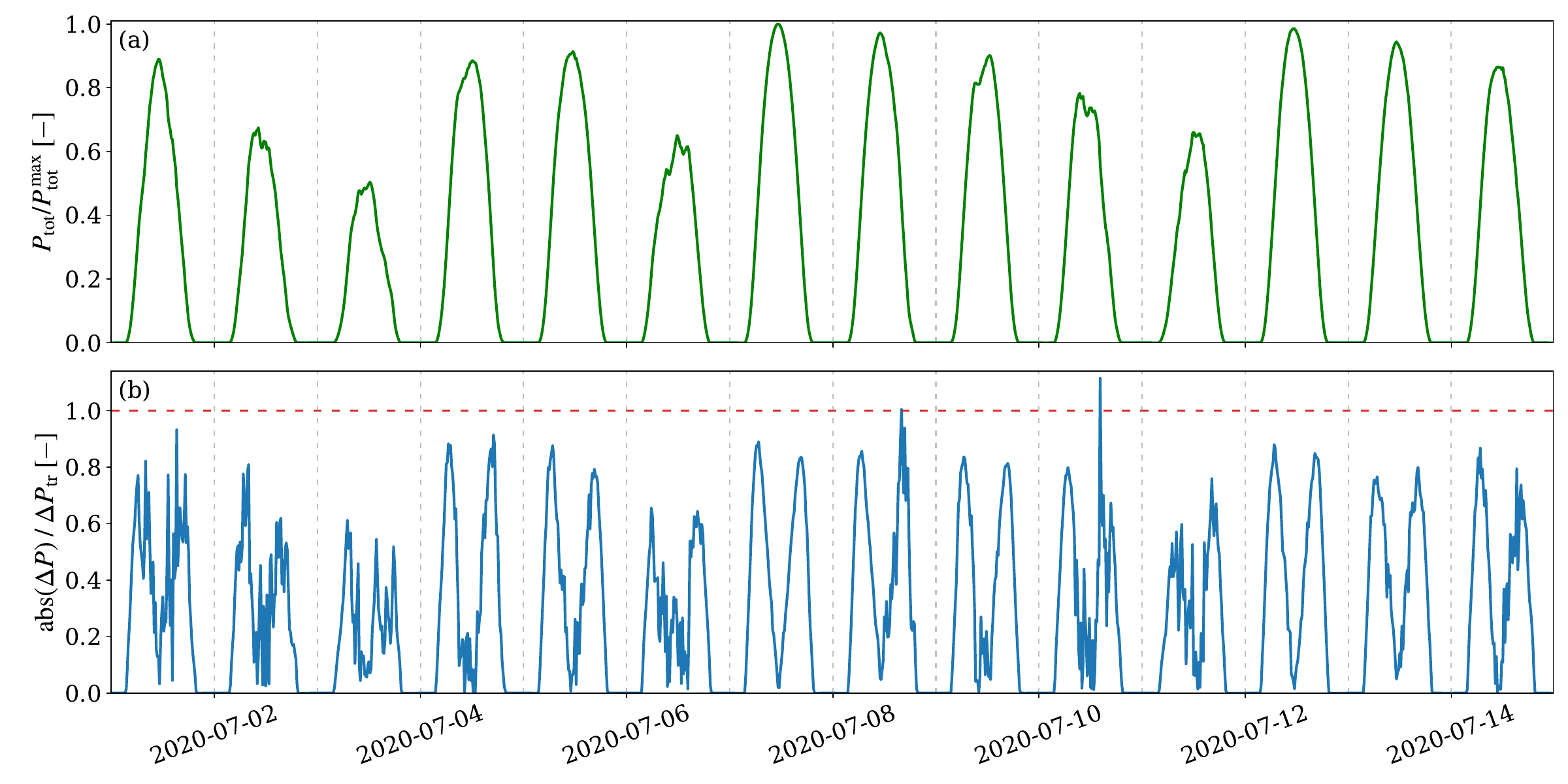}
    \caption{(a) Time series of aggregated PV power production from 6434 PV stations, normalized by the maximum value observed over the displayed period, and (b) the corresponding magnitude of power variations, normalized by the ramp threshold $\Delta P_{\mathrm{tr}}$. Results are shown for the first two weeks of July~2020. The horizontal dashed red line indicates the threshold used to identify ramp events, while the vertical dashed grey lines mark midnight of each day.}
	\label{fig:power_and_ramp}
\end{figure}

\section{Solar ramp events}\label{sec:ramp_events}
Ramp events do not have a universally accepted definition, as their characterization depends on the specific application and operational context. Ramp identification relies on several key attributes, including the duration $\Delta t$, magnitude of the power variation $\Delta P$, rate of change $\Delta P / \Delta t$, and the direction of the transition \cite{Wang2022}. Ramp direction distinguishes between increases and decreases in power output. Positive ramps (ramp-up events) correspond to periods of rapidly increasing generation, whereas negative ramps (ramp-down events) denote sharp decreases in power production. The severity of a ramp event is typically determined by the interplay between its magnitude and duration. Large power variations occurring over short time intervals are generally considered more critical, as they imply higher rates of change and pose greater operational challenges for grid management.

\begin{figure}[t]
	\centering
	\includegraphics[width=1\textwidth]{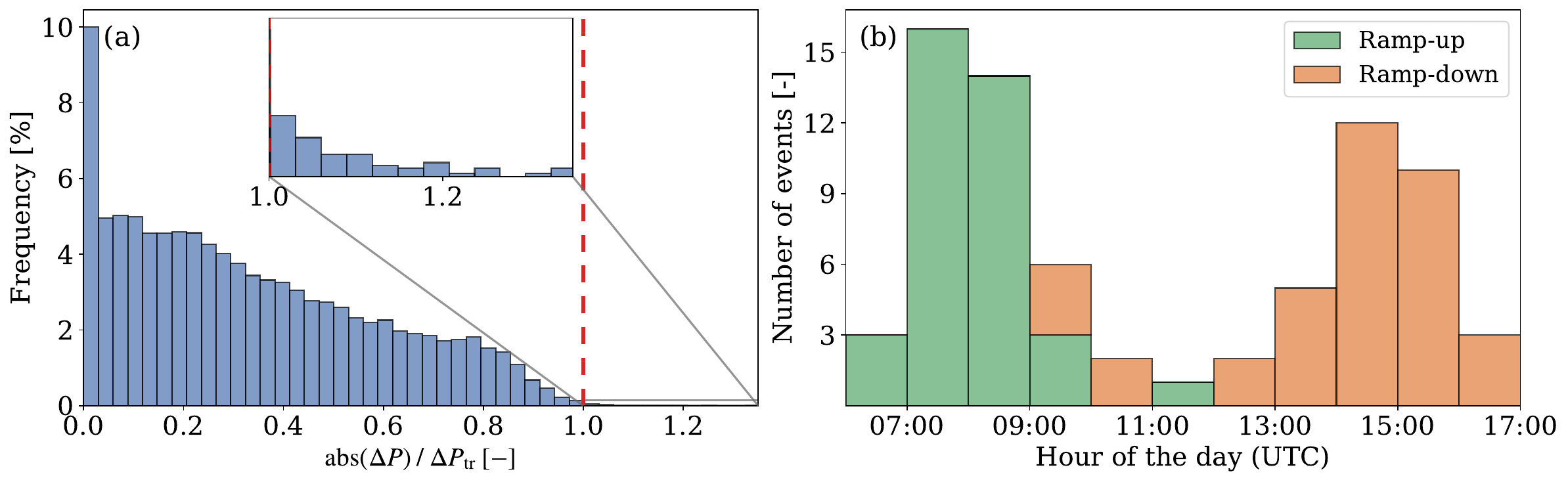}
	\caption{(a) Empirical probability density of magnitude of the power variation over a time interval of 15 minutes normalized by the threshold $\Delta P_\mathrm{tr}$, measured over the year 2019-2020 for the aggregate PV power generation of 6434 PV stations. (b) Number of ramp events aggregated by hour of the day, with distinction between ramp-up and ramp-down events.}
	\label{fig:ramp_statistics_combined}
\end{figure}

In this study, the ramp duration is fixed to the temporal resolution of both the SSI and PV datasets, that is, 15 minutes. This choice is consistent with the time granularity adopted nowadays in electricity market operations \cite{Markle2018}. The primary challenge lies in defining an appropriate threshold for $\Delta P$ to distinguish ramp events from normal variability. To address this, we adopt a data-driven approach. First, satellite observations are used to compute the spatially averaged CSI over the study domain between sunrise and sunset for each day within the two-year study period. From this distribution, we identify a subset of 43 days characterised by predominantly clear-sky conditions, defined as those with a mean CSI between sunrise and sunset exceeding the 90th percentile. Next, we aggregate the PV power production of all 6434 PV systems into a single country-level generation time series and compute the maximum 15-minute power variation observed during these clear-sky days. This value represents the upper bound of power fluctuations in clear-sky days, i.e., in the absence of cloud-induced variability and solely driven by deterministic diurnal evolution. The resulting threshold, $\Delta P_\mathrm{tr}=37.7$ MW, is adopted to differentiate between non-ramp and ramp events. This corresponds to a rate of change of 2.51 MW min$^{-1}$. Power variations exceeding $\Delta P_\mathrm{tr}$ over a 15-minute interval are therefore classified as ramp events, as they exceed the variability expected under clear-sky conditions and are attributed to transient atmospheric phenomena such as cloud dynamics. We note that this threshold corresponds to approximately 7.1\% of the aggregated $P_{95}$ values of all PV systems considered. Importantly, this threshold is dataset-specific and depends on both the spatial aggregation level and the chosen temporal resolution, and therefore must be determined for the particular set of monitored PV systems. 

\begin{figure}[t]
	\centering
	\includegraphics[width=1\textwidth]{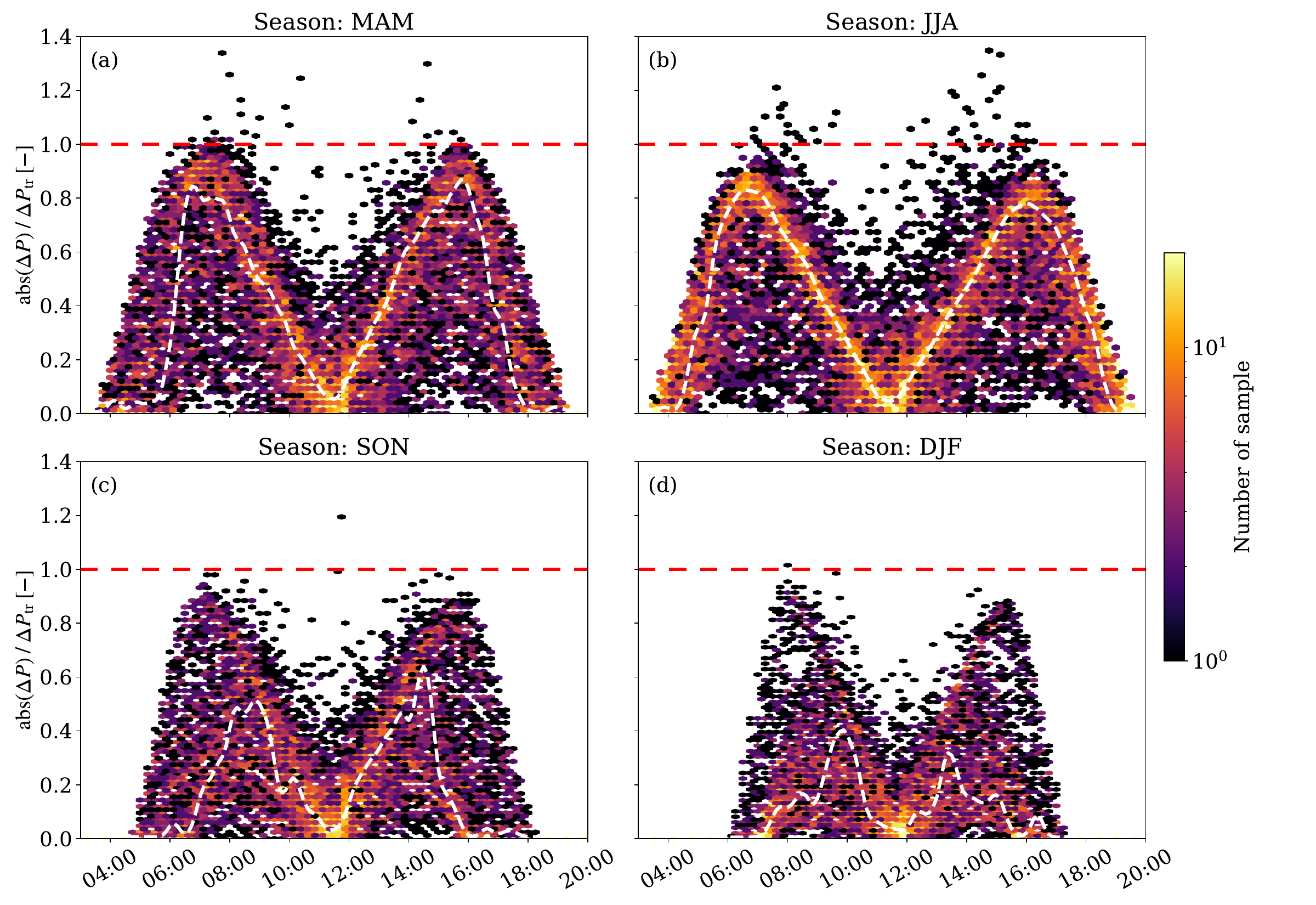}
	\caption{Hexagonal bin distribution of ramp magnitude normalized with the threshold $\Delta P_\mathrm{tr}$ as a function of hour of the day for (a) MAM, (b) JJA, (c) SON, and (d) DJF. Color shading indicates the number of samples per hexagonal bin on a logarithmic scale. The dashed white line traces the locus of maximum sample density at each hour, highlighting the most probable ramp magnitude throughout the diurnal cycle. The dashed red line marks the threshold used to identify ramp events. The bin widths are 15 minutes along the x-axis and 0.028 in normalized ramp units along the y-axis.}
	\label{fig:ramp_per_season_hexbin}
\end{figure}

The aggregated PV power generation of the 6434 PV stations considered in this study is shown in Figure \ref{fig:power_and_ramp}(a) for the first two weeks of July 2020. The total PV power output exhibits a pronounced diurnal cycle, with day-to-day differences in peak production driven mainly by varying cloud conditions. Periods of reduced peak power indicate the presence of extended cloud cover, while smoother and higher daytime maxima correspond to predominantly clear-sky conditions. The associated power variations over the same time period are shown in Figure \ref{fig:power_and_ramp}(b). High values of $|\Delta P|/\Delta P_{\mathrm{tr}}$ occur primarily during morning and afternoon in clear-sky days, whereas rapid fluctuations in PV power output on cloudy or highly variable days are driven by transient cloud passages. Conversely, low values indicate stable production regimes under slowly varying irradiance conditions. Power variations exceeding the defined threshold (red dashed line) identify ramp events, highlighting time intervals with extreme short-term variability.

The distribution of the 15-minute power variations computed from the aggregated PV generation over the two-year time span is shown in Figure~\ref{fig:ramp_statistics_combined}(a). The distribution is strongly skewed toward small amplitudes, indicating that most power variations remain well below the $\Delta P_{\mathrm{tr}}$ threshold. The frequency decreases nearly monotonically with increasing $\Delta P$. In total, we identify 74 events with $\Delta P > \Delta P_{\mathrm{tr}}$. We refer to these as ramp events, which may be considered extreme because they impose relevant stress on the power grid. These events reside in the upper tail of the distribution, highlighting the comparatively low probability of large-amplitude ramps. Due to their low frequency, such events are inherently challenging for machine learning models to represent accurately \cite{Olivetti2024}. Figure~\ref{fig:ramp_statistics_combined}(b) shows the diurnal distribution of the 74 ramp events. Ramp-up events occur predominantly during the morning (approximately 07:00–10:00 UTC), whereas ramp-down events are concentrated in the afternoon (approximately 13:00–16:00 UTC). This asymmetry reflects systematic differences in the physical drivers of strong positive and negative PV power changes. As will be discussed later, ramp-up events are typically associated with cloud dissipation during the morning, while ramp-down events commonly occur when cloud cover increases in the afternoon. In fact, these processes reinforce the underlying diurnal cycle of clear-sky SSI, amplifying the increase in solar irradiance in the morning and the decrease in the afternoon.

The seasonal distribution of aggregated PV power variations as a function of the time of day using hexagonal binning is illustrated in Figure~\ref{fig:ramp_per_season_hexbin}. The color scale indicates the number of samples within each hexagonal bin, thereby visualizing the density of occurrences. Across all seasons, most samples lie below the threshold $\Delta P_{\mathrm{tr}}$, confirming that ramp events are rare and confined to the upper tail of the distribution. A pronounced diurnal structure is evident in every season. Large power variations are clustered around the morning and afternoon hours, whereas variations near solar noon are typically smaller, reflecting comparatively stable generation conditions. The dashed white line indicates the regions of highest occurrence density. The latter mostly follow the derivative of the diurnal evolution of SSI in clear-sky days, where the largest gradients occur during the morning rise and afternoon decline of the sun in the sky. Seasonal differences are clearly visible. In spring (MAM) and summer (JJA), the distribution spans a broader temporal range, consistent with longer daylight duration and higher irradiance levels. In autumn (SON), the temporal window narrows and the density shifts toward moderately lower magnitudes. During winter (DJF), both the occurrence density and the amplitude of power variations are substantially reduced, and events are confined to a short midday interval, reflecting limited daylight hours and lower solar elevation. Ramp events occur predominantly in the MAM and JJA months, and are most frequent during the morning and afternoon hours, consistent with the statistics shown in Figure~\ref{fig:ramp_statistics_combined}(b).

\begin{figure}[t]
	\centering
	\includegraphics[width=1\textwidth]{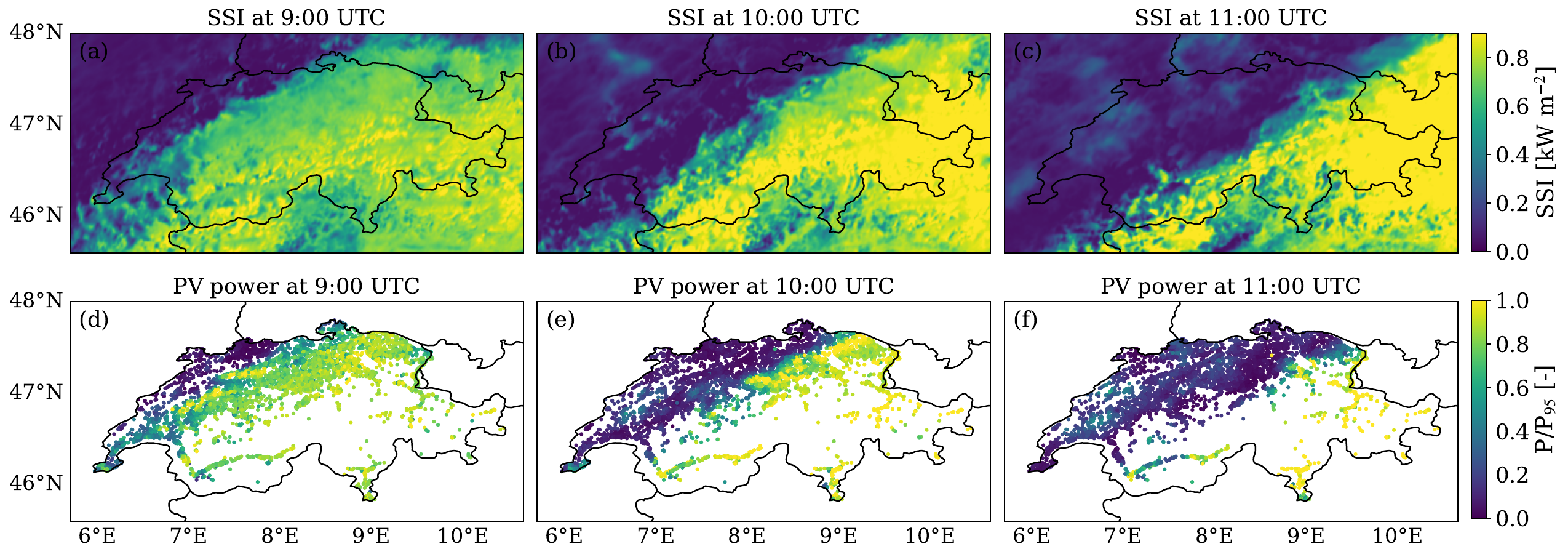}
	\caption{(a-c) Satellite-based SSI fields and (d-f) PV power output of the 6434 PV stations, observed on 23 May 2020 at three time stamps: 09:00 UTC, 10:00 UTC and 11:00 UTC. The black lines denote national borders. Note that the PV power output is normalized using the station-specific 95th percentile of the power time series.}
	\label{fig:ssi_and_power_ramp_23_05_2020}
\end{figure}

Finally, we present an example of cloud dynamics leading to a ramp event. Specifically, Figure~\ref{fig:ssi_and_power_ramp_23_05_2020}(a–c) shows the satellite-derived SSI fields at 09:00, 10:00, and 11:00 UTC, while Figure~\ref{fig:ssi_and_power_ramp_23_05_2020}(d–f) presents the corresponding PV power production at the same times. At 09:00 UTC, high SSI values prevail over large parts of the region of interest, consistent with elevated PV power production across most stations. Between 09:00 and 11:00 UTC, a coherent cloud front is advected to the southeast, covering the majority of the Swiss Plateau and causing a significant reduction in SSI. The distributed PV power production reveals a corresponding spatially coherent decline in PV power generation in the areas affected by the cloud band. The temporal sequence highlights the strong coupling between satellite-observed irradiance variability and the response of distributed PV systems. This event illustrates how the passage of a cloud band over a region with a high density of PV installations can induce a rapid decline in PV power production. In fact, between 09:00 and 11:00 UTC, the national PV power output decreased by approximately 60.7\%.

\section{Methodology}\label{sec:methodology}
The satellite-based nowcasting models, namely IrradianceNet, SolarSTEPS, and SHADECast, are adopted to forecast satellite-derived CSI fields. All models were originally trained and calibrated on seven years of HelioMont CSI data at a spatial resolution of 0.02$^\circ$ -- see Section \ref{sec:data} for more information. For evaluation, we use two years of HANNA SSI data at 0.01$^\circ$ resolution. To ensure consistency with the training setup, the HANNA SSI fields are downsampled to 0.02$^\circ$ using 2×2 pixel averaging. SSI is subsequently converted to CSI using clear-sky irradiance computed with the Ineichen model as implemented in pvlib \cite{Ineichen2008, pvlib2018}. Each forecast is generated from four consecutive CSI fields (corresponding to one hour of observations) and predicts the subsequent eight time steps, corresponding to a two-hour forecast horizon with lead times of 15 minutes. The predicted CSI fields are finally converted back to SSI, resulting in the corresponding SSI forecasts.

SHADECast and IrradianceNet operate over the spatial domain [4.41$^\circ$E, 12.09$^\circ$E] × [42.99$^\circ$N, 50.67$^\circ$N], corresponding to images of size 384 × 384 pixels. Due to architectural constraints, IrradianceNet is applied using a 128 × 128 patch-based strategy, with border interpolation employed to reconstruct the full spatial field \cite{Carpentieri2025, Nielsen2021}. SolarSTEPS requires a slightly larger input domain to mitigate edge effects associated with optical-flow-based advection. The predicted irradiance fields are subsequently converted into PV power using station-specific machine learning models, enabling comparison with production data from 6434 PV stations across the study domain. We adopt the extreme gradient boosting (XGBoost) algorithm \cite{Chen2016} to model the conversion between solar irradiance and PV power output. The SSI serves as the primary meteorological predictor. To account for geometric and topographic effects, such as terrain-induced shading, we additionally include the solar zenith angle (SZA) and solar azimuth angle (AZI). These quantities are computed using HORAYZON, a ray-tracing algorithm that determines the local horizon and sky-view factor \cite{Steger2022}. This is particularly important in regions with complex topography, where terrain features can induce substantial spatial variability in SSI. To capture temporal periodicity, we further incorporate the day of the year (DoY) and the hour of the day (HoD). In total, each PV system is characterized by seven predictors, i.e. SSI, SZA, AZI, and the four time-based features (i.e., HoD and DoY with cyclical encoding), and one target variable, namely the measured PV power output. To account for differences in nominal capacity, orientation, and elevation among PV systems, a separate XGBoost model is trained for each station. We refer to \cite{Lanzilao2026} for further details on the training strategy and model performance.

Forecast evaluation follows a user-centric setup. At each full hour, between one hour after sunrise and three hours before sunset, the models generate forecasts with a two-hour horizon and time resolution of 15 minutes. Although the satellite-based models are capable of producing forecasts at 15-minute intervals, the evaluation is restricted to hourly initialization to ensure comparability with IFS-ENS. This procedure is repeated over the entire two-year evaluation period, yielding a total of 6158 forecast inferences per model. We note that the methodology used for PV power forecasting corresponds to that implemented by \cite{Lanzilao2026}, to which we refer for a detailed description. All simulations were performed on the Swiss high-performance computing system Alps at the Swiss National Supercomputing Centre.

\section{Results}\label{sec:results}
The aggregate performance of the forecasting models across all ramp events is presented in Section \ref{sec:results_1}, along with a comparison against non-ramp conditions using both deterministic and probabilistic metrics. Subsequently, three case studies are selected to illustrate model performance under typical weather regimes associated with ramp events. Section \ref{sec:results_2} examines a ramp-down event driven by cloud advection, Section \ref{sec:results_3} presents a ramp-up event resulting from rapid morning cloud dissipation, and Section \ref{sec:results_4} highlights how fog and low-level clouds dissipation over the Swiss Plateau can lead to strong increases in PV power generation. A more detailed quantitative analysis of these case studies is provided in \ref{app:appendix_a}.

\begin{figure}[t]
	\centering
	\includegraphics[width=1\textwidth]{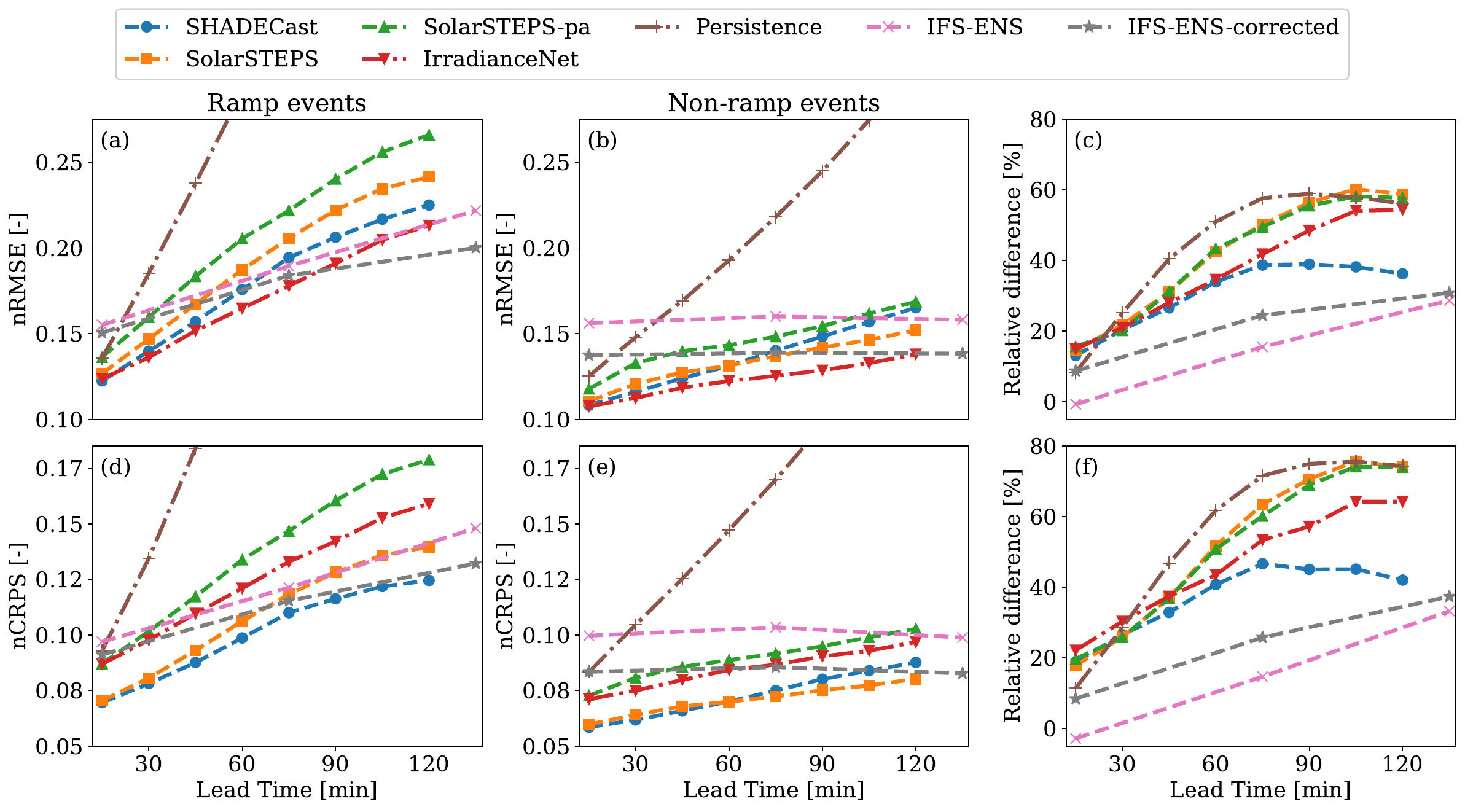}
	\caption{(a,b) nRMSE and (d,e) nCRPS averaged across all PV stations and over the subsets of forecasts corresponding to (a,d) ramp events and (b,e) non-ramp events. Panels (c,f) show the relative difference in nRMSE and nCRPS, respectively, between forecasts during ramp and non-ramp conditions. The relative difference is defined as $(m_A - m_B) / m_B$, where $m$ denotes a generic metric and $A$ and $B$ refer to ramp and non-ramp subsets, respectively. For deterministic models, nCRPS is replaced by nMAE. Dashed–dotted lines indicate deterministic models, while dashed lines represent probabilistic models.}
	\label{fig:power_comparison_line_ramp_yes}
\end{figure}

\begin{figure}[t]
	\centering
	\includegraphics[width=1\textwidth]{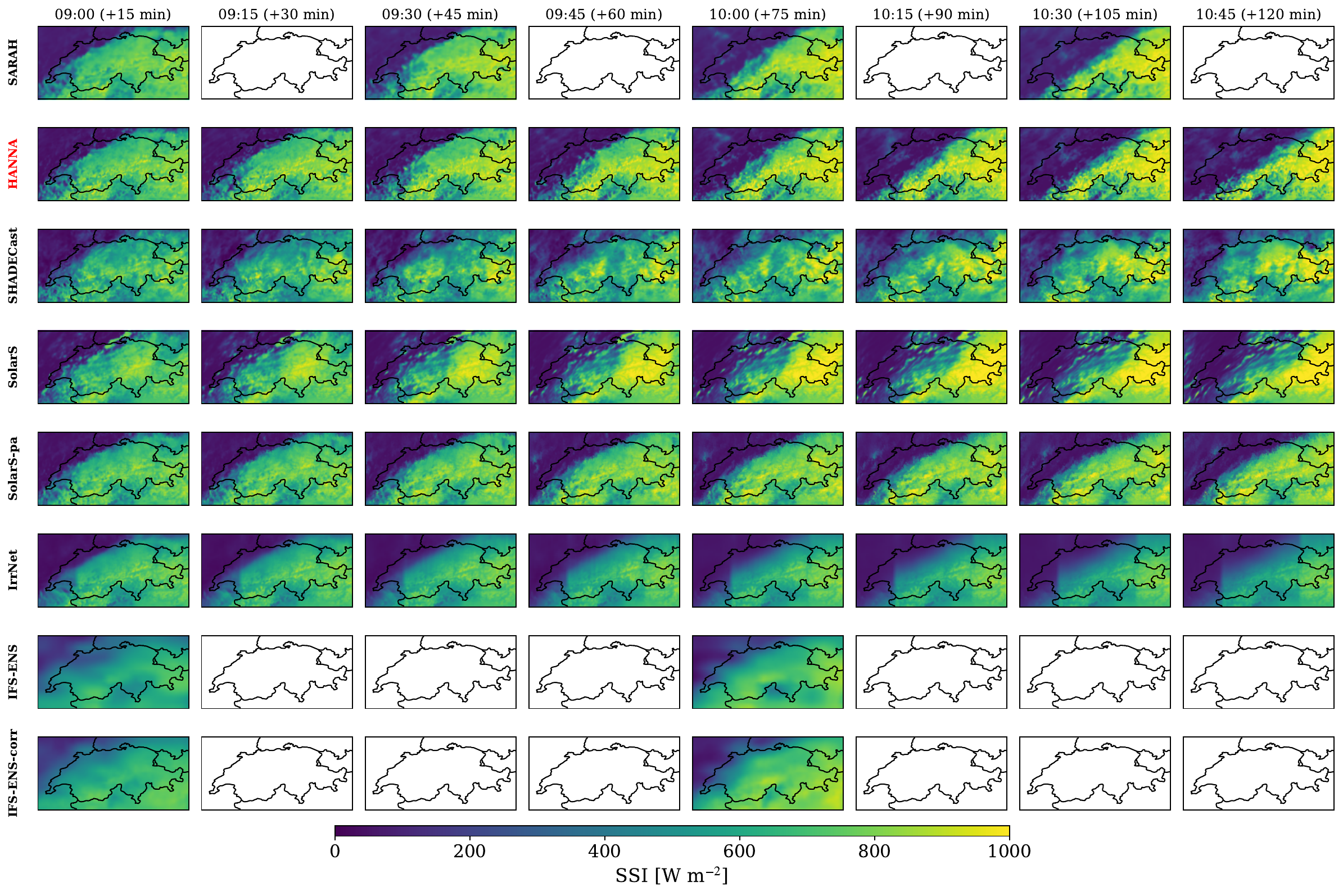}
	\caption{Satellite-based SSI observations and model forecasts over the area of interest at lead times ranging from 15 to 120 minutes. The forecasts are issued at 09:00 UTC on 23 May 2020. For the probabilistic models, the ensemble member chosen is the one with the lowest RMSE. Note that the IFS-ENS SSI fields represent a one-hour temporal average, whereas all other fields are instantaneous. The black lines denote national borders. The row highlighted with the red label is used as the ground truth. Missing panels reflect differences in the temporal resolution of the satellite observations and forecast models. For illustration purposes, the SolarSTEPS, SolarSTEPS-pa and IrradianceNet models are abbreviated to SolarS, SolarS-pa and IrrNet, respectively.}
	\label{fig:comparison_inference_ssi_1}
\end{figure}

\begin{figure}[t]
	\centering
	\includegraphics[width=1\textwidth]{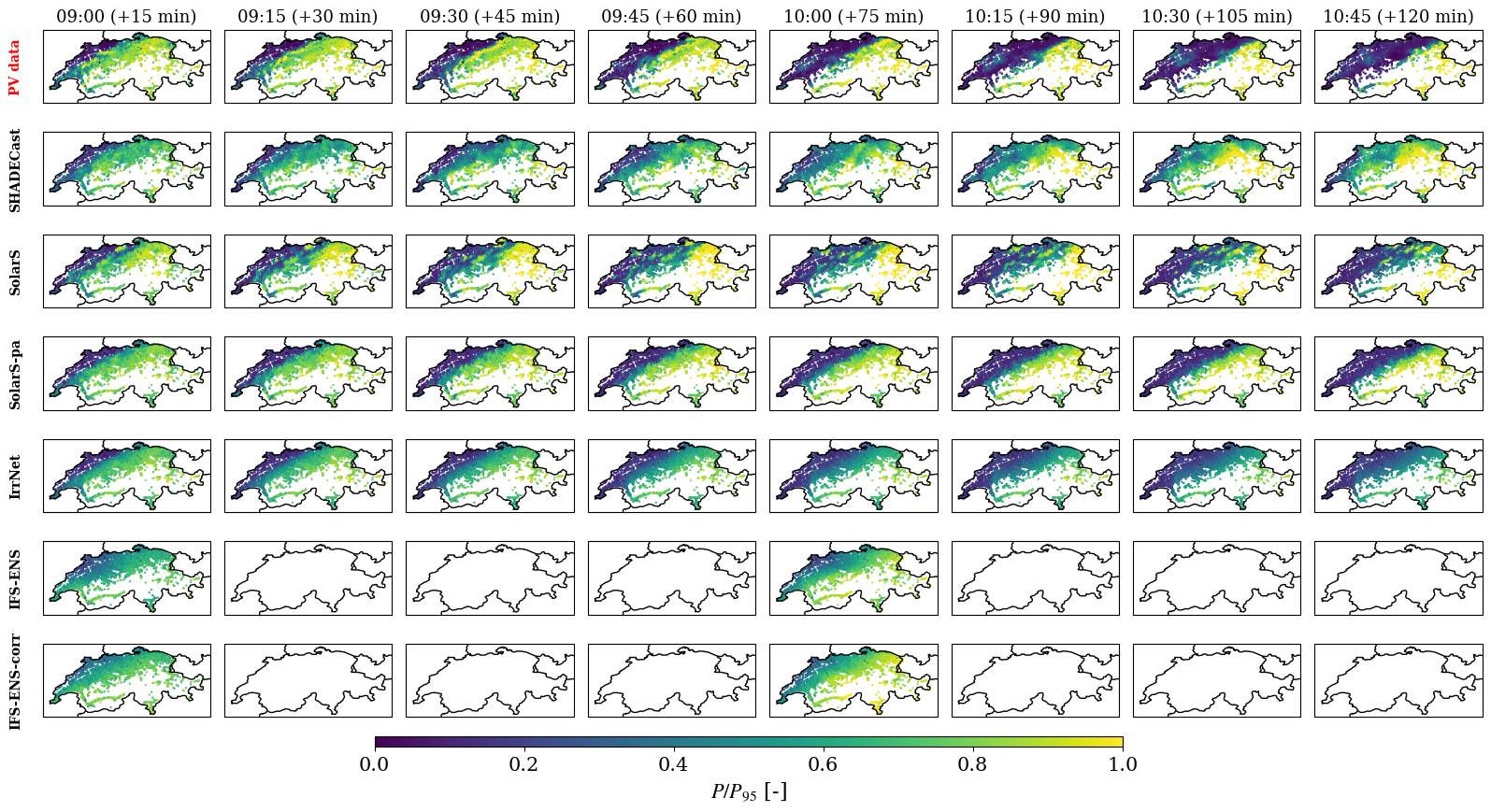}
	\caption{PV power measurements and model forecasts at lead times ranging from 15 to 120 minutes. The forecasts are issued at 09:00 UTC on 23 May 2020. For probabilistic models, the same ensemble member used in Figure \ref{fig:comparison_inference_ssi_1} is shown. The row highlighted with the red label is used as the ground truth. Missing panels reflect differences in the temporal resolution of the satellite observations and forecast models. Note that the PV power output is normalized using the station-specific 95th percentile of the power time series.}
	\label{fig:comparison_inference_power_1}
\end{figure}

\subsection{Statistical evaluation across all ramp events}\label{sec:results_1}
To assess the performance of the models in predicting upcoming ramps, we identify all forecasts that contain at least one ramp event within the 30–75 minute lead-time window. This window is chosen to focus on short-term predictability while avoiding very short lead times, where persistence effects dominate, and longer lead times, where forecast skill typically decreases. As 29\% of ramp events occur at consecutive time steps, this results in a total of 51 forecasts classified as ramp-event cases. Notably, these represent only 0.83\% of all forecasts over the 2019–2020 period, highlighting the rarity and extreme nature of such events. To evaluate model performance under contrasting conditions, we construct a corresponding set of forecasts without ramp events. For each ramp-event case, we select forecasts issued at the same UTC time on the preceding and following days, provided that no ramp occurs within the same lead-time window. This approach ensures a consistent comparison framework, as the selected forecasts are temporally close and therefore associated with similar solar elevation angles and typical SSI levels. Consequently, differences in forecast performance can be more directly attributed to the presence or absence of ramp dynamics rather than to seasonal or diurnal variability. Applying this procedure yields a total of 77 forecasts classified as non-ramp cases. Forecast performance is evaluated using both deterministic and probabilistic metrics. The normalized root mean square error (nRMSE) quantifies the magnitude of point forecast errors, representing the average deviation between predicted and observed power. The normalized continuous ranked probability score (nCRPS) assesses the full predictive distribution, capturing both reliability and sharpness of the ensemble forecasts. Detailed definitions of these metrics are provided in \ref{app:appendix_b}.

Figure \ref{fig:power_comparison_line_ramp_yes} compares PV power forecast errors, averaged over the ramp and non-ramp subsets, as a function of lead time. For both deterministic (nRMSE) and probabilistic (nCRPS) metrics, errors increase with lead time across all models. However, the degradation in forecast skill is substantially more pronounced during ramp events, highlighting the difficulty of predicting PV power under rapidly evolving cloud conditions. The optical-flow-based approach that relies solely on advection (i.e. SolarSTEPS-pa) exhibits the strongest error growth in ramp situations, reflecting that the model does not account for cloud formation and dissipation processes. Among the satellite-based models, SHADECast emerges as the most reliable, achieving an nCRPS that is 10.8\% lower than that of SolarSTEPS at a 120-minute lead time. Under non-ramp conditions, however, SolarSTEPS slightly outperforms SHADECast at lead times beyond 60 minutes, consistent with the findings of \cite{Lanzilao2026}. As expected, the bias-corrected IFS forecasts consistently outperform the raw forecasts, highlighting the importance of statistical post-processing. Figure \ref{fig:power_comparison_line_ramp_yes}(c,f) further quantifies the relative increase in forecast error between ramp and non-ramp cases in terms of nRMSE and nCRPS, respectively. For most satellite-based models, ramp events lead to nRMSE increases of 20–30\% at short lead times, rising to approximately 60\% at longer horizons. A similar pattern is observed for the nCRPS, with relative differences reaching up to 70\% beyond 90-minute lead times. Although IFS-ENS also shows performance degradation during ramp events, the relative increase is less pronounced compared to the satellite-based models. Overall, the results demonstrate that ramp conditions not only reduce point forecast accuracy but also deteriorate the probabilistic skill of ensemble predictions, making ramp events a dominant source of forecast uncertainty in PV power forecasting.

\begin{figure}[t]
	\centering
	\includegraphics[width=1\textwidth]{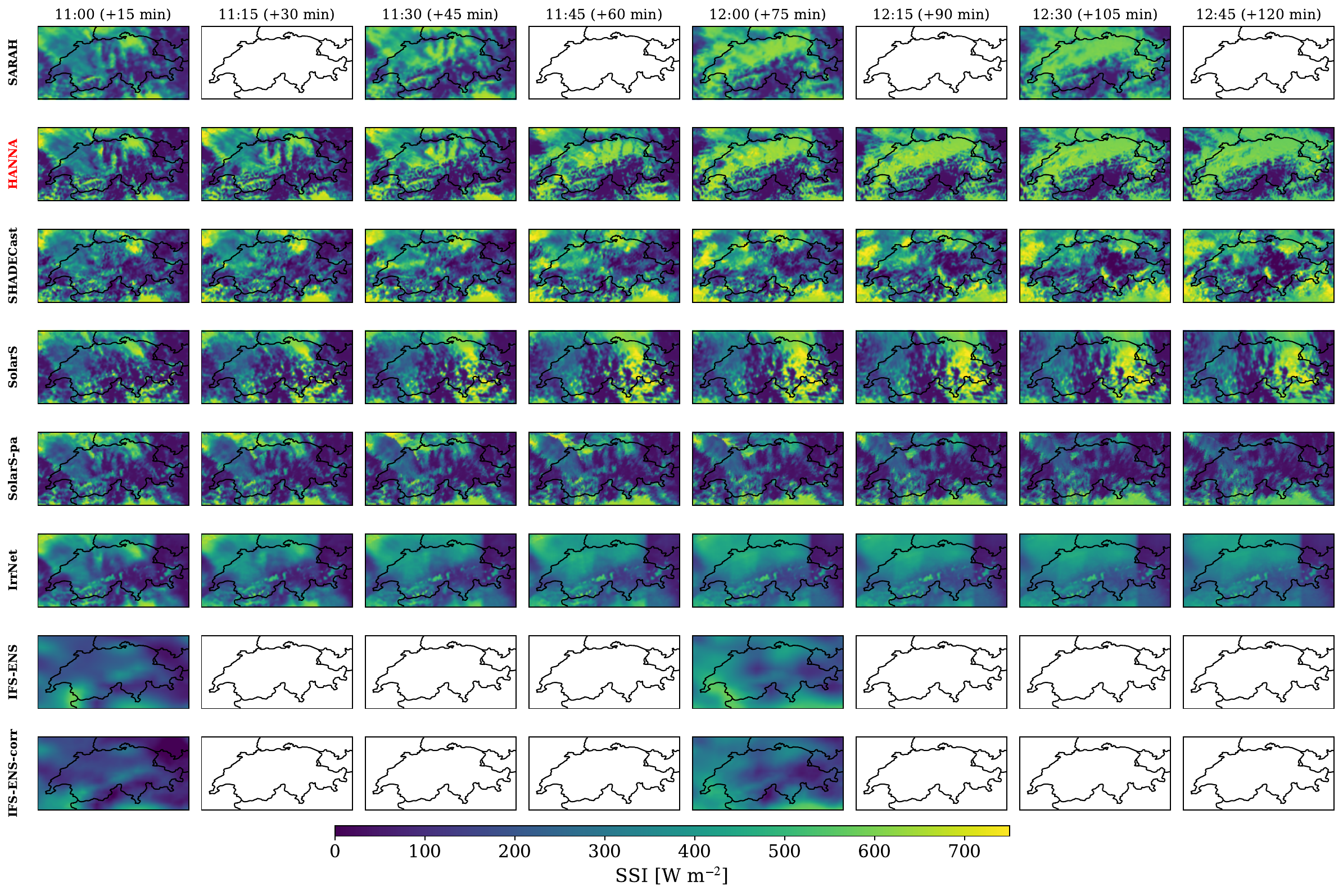}
	\caption{Satellite-based SSI observations and model forecasts over the area of interest at lead times ranging from 15 to 120 minutes. The forecasts are issued at 11:00 UTC on 3 October 2020. For the probabilistic models, the ensemble member chosen is the one with the lowest RMSE. Note that the IFS-ENS SSI fields represent a one-hour temporal average, whereas all other fields are instantaneous. The black lines denote national borders. The row highlighted with the red label is used as the ground truth. Missing panels reflect differences in the temporal resolution of the satellite observations and forecast models.}
	\label{fig:comparison_inference_ssi_2}
\end{figure}

\begin{figure}[t]
	\centering
	\includegraphics[width=1\textwidth]{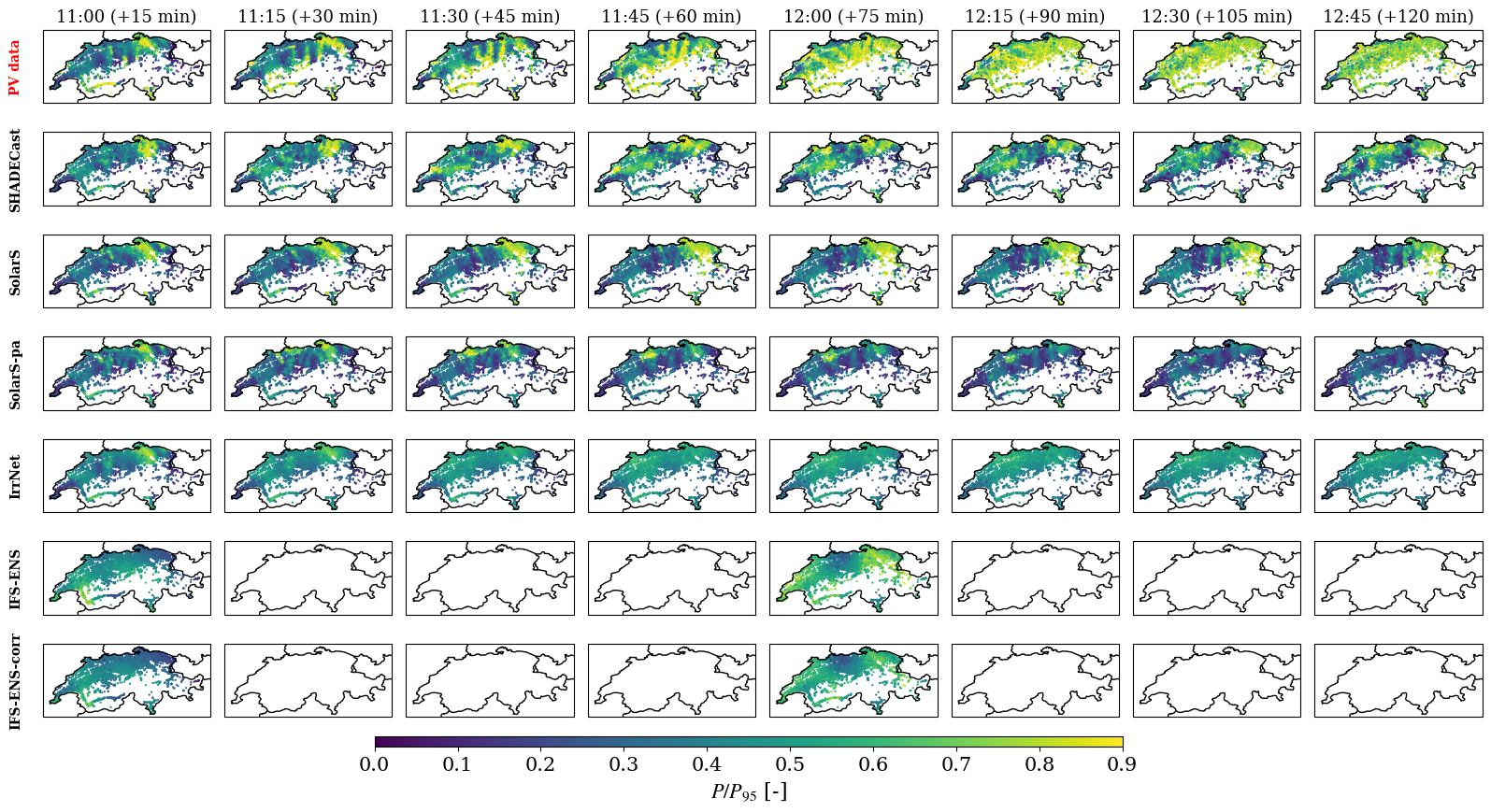}
	\caption{PV power measurements and model forecasts at lead times ranging from 15 to 120 minutes. The forecasts are issued at 11:00 UTC on 3 October 2020, a day with high-variability weather. For probabilistic models, the same ensemble member used in Figure \ref{fig:comparison_inference_ssi_2} is shown. The black lines denote national borders. The row highlighted with the red label is used as the ground truth. Missing panels reflect differences in the temporal resolution of the satellite observations and forecast models. Note that the PV power output is normalized using the station-specific 95th percentile of the power time series.}
	\label{fig:comparison_inference_power_2}
\end{figure}

\begin{figure}[t]
	\centering
	\includegraphics[width=1\textwidth]{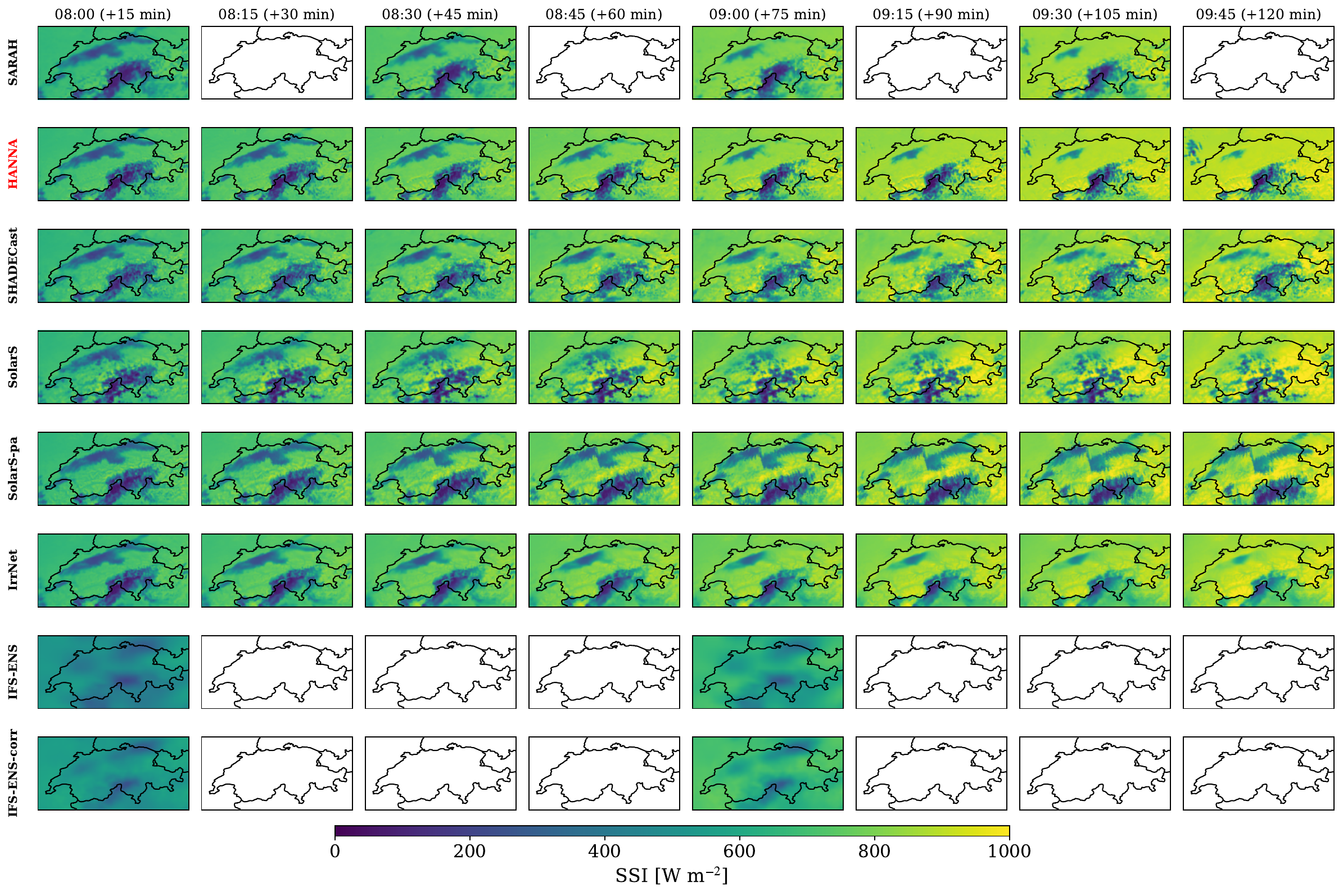}
	\caption{Satellite-based SSI observations and model forecasts over the area of interest at lead times ranging from 15 to 120 minutes. The forecasts are issued at 08:00 UTC on 7 June 2019. For the probabilistic models, the ensemble member chosen is the one with the lowest RMSE. Note that the IFS-ENS SSI fields represent a one-hour temporal average, whereas all other fields are instantaneous. The black lines denote national borders. The row highlighted with the red label is used as the ground truth. Missing panels reflect differences in the temporal resolution of the satellite observations and forecast models.}
	\label{fig:comparison_inference_ssi_3}
\end{figure}

\begin{figure}[t]
	\centering
	\includegraphics[width=1\textwidth]{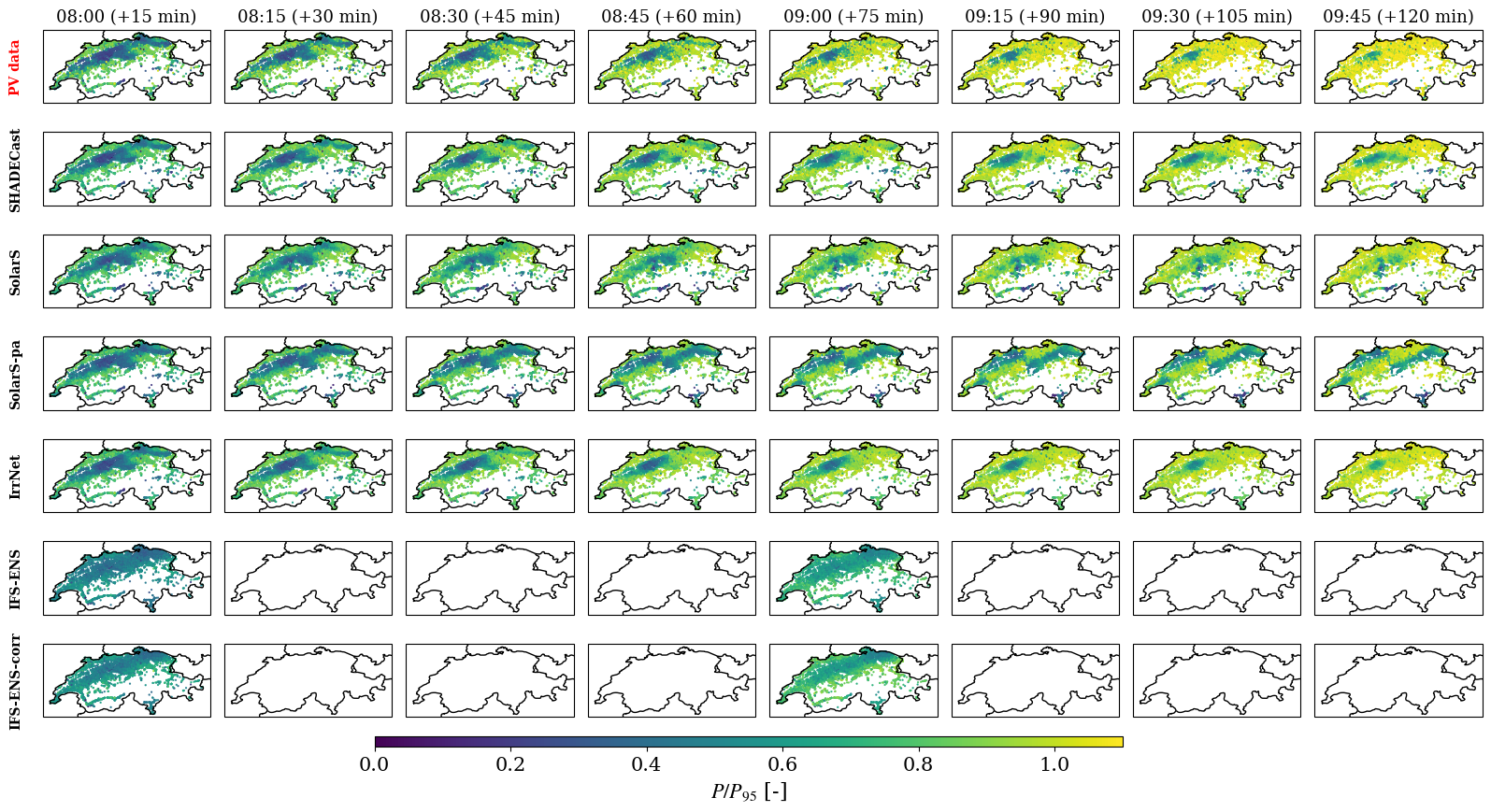}
	\caption{PV power measurements and model forecasts at lead times ranging from 15 to 120 minutes. The forecasts are issued at 08:00 UTC on 7 June 2019. For probabilistic models, the same ensemble member used in Figure \ref{fig:comparison_inference_ssi_3} is shown. The black lines denote national borders. The row highlighted with the red label is used as the ground truth. Missing panels reflect differences in the temporal resolution of the satellite observations and forecast models. Note that the PV power output is normalized using the station-specific 95th percentile of the power time series.}
	\label{fig:comparison_inference_power_3}
\end{figure}

\subsection{Case study I: ramp-down event driven by cloud advection}\label{sec:results_2}
The SSI satellite observations and model forecasts for the ramp-down event described in Section \ref{sec:ramp_events} are presented in Figure \ref{fig:comparison_inference_ssi_1} for lead times ranging from 15 to 120 minutes. Forecasts are issued at 09:00 UTC on 23 May 2020, using satellite-observed SSI fields from 08:00 to 08:45 UTC as input. The SSI observations are shown for two retrieval methods, that is SARAH-3 and HANNA. Both datasets show a coherent cloud band that is advected across the study domain in a south-easterly direction. Due to its higher spatial and temporal resolution, HANNA more clearly resolves cloud edges, small-scale cloud structures, and their temporal evolution. This ramp-down event is primarily governed by cloud advection. Consequently, the optical-flow-based models, SolarSTEPS and SolarSTEPS-pa, succeed in forecasting the cloud band motion with relatively high accuracy. While SHADECast captures the SSI increase in eastern Switzerland, it fails to accurately predict the cloud-band advection in this ramp event, likely due to the limited representation of such events in the training data. At the same time, the absence of spatial blurring, attributable to its diffusion-based framework, is evident. The IrradianceNet model captures the overall evolution of the cloud band better but fails to reproduce the increase in SSI over the eastern regions. Moreover, the patch-based strategy introduces spatial discontinuities particularly evident at lead times exceeding 60 minutes, with sharp transitions along patch boundaries where interpolation is applied. In addition, its forecasts are smoother and highly diffuse. Finally, IFS-ENS correctly predicts regions of low SSI. However, its comparatively coarse spatial and temporal resolution yields smooth and more homogeneous SSI fields with less distinct cloud boundaries.

Figure \ref{fig:comparison_inference_power_1} presents the corresponding PV power forecasts obtained from the station-specific machine-learning-based irradiance-to-power conversion model. The top row shows the observed PV power generation, which exhibits pronounced spatial variability closely aligned with the cloud distribution. As the cloud band advects over the Swiss Plateau, national PV power output decreases by approximately 60.7\% between 09:00 and 10:45 UTC. Overall, the forecast models reproduce the main SSI patterns, with low-irradiance areas corresponding to reduced PV power output and vice versa. Given that this ramp event is dominated by cloud advection, the displayed ensemble member of the SolarSTEPS-pa forecast is in close agreement with the observed PV power distribution, while SHADECast struggles to accurately capture the power gradients in this case.

\subsection{Case study II: ramp-up event driven by cloud dissipation}\label{sec:results_3}
The second illustrative case highlights a ramp-up event that occurred on 3 October 2020 at 11:45 UTC. Figure \ref{fig:comparison_inference_ssi_2} presents the SSI satellite observations and model forecasts for the ramp-up event described in Section \ref{sec:ramp_events}, also for lead times between 15 and 120 minutes. Forecasts were initialized at 11:00 UTC on 3 October 2020, using SSI fields from 10:00 to 10:45 UTC. Both SARAH and HANNA retrievals indicate rapid cloud clearing over the Swiss Plateau. This large-scale cloud dissipation leads to a significant increase in SSI and, consequently, a strong ramp-up in PV power production. In contrast to the previously discussed advection-dominated case, cloud dissipation plays a major role in this event, posing a challenge for the optical-flow-based models. In particular, SolarSTEPS-pa is unable to represent the rapid reduction in cloud cover and therefore substantially underestimates the irradiance increase. While showing tendencies to increasing SSI in the forecasts, the data-driven models also struggle to reproduce the rapid clearing, likely reflecting the limited representation of such events in the training data. The IFS-ENS shows a comparable behaviour, with a smoothed transition toward higher SSI values. As a result, all models underestimate both the magnitude and spatial extent of the ramp-up. 

Figure \ref{fig:comparison_inference_power_2} displays the corresponding PV power forecasts. The observed power production increases by approximately 65.7\% between 10:00 and 11:45, in line with the clearing of clouds. While the models broadly reflect the relationship between irradiance and power, they fail to capture the steep spatial gradients and rapid temporal increase associated with this event. Overall, these results highlight that rapid changes in cloud cover remain difficult to predict, regardless of whether the forecasting approach is data-driven or physics-based, remaining a major source of forecast error in both SSI and PV power predictions.

\subsection{Case study III: Ramp-up event in presence of low-level clouds}\label{sec:results_4}
The Swiss Plateau is frequently affected by fog and low-level clouds due to orographic confinement and the abundant moisture supply from lakes and rivers. During winter, such stratiform cloud decks often persist throughout the day under stable atmospheric conditions. In late spring and early summer, low-level clouds typically dissipate during the morning as solar heating intensifies quickly. Because a large fraction of PV installations are located in the Swiss Plateau, this transition from fog and overcast to clear-sky conditions, combined with the increasing solar elevation, frequently triggers large variations in PV power output. One exemplary case occurred on 7 June 2019 at 08:45 UTC. The corresponding SSI satellite observations and model forecasts for lead times from 15 to 120 minutes are shown in Figure \ref{fig:comparison_inference_ssi_3}. At 08:00 UTC, the satellite observations indicate a widespread layer of clouds covering the majority of the Swiss Plateau. As the sun rises, enhanced surface heating warms the boundary layer, increases turbulent mixing, and progressively dissipates the cloud deck. By 10:00 UTC, most clouds have dissipated and the region transitions to near clear-sky conditions. Data-driven models, such as SHADECast and IrradianceNet, reproduce the cloud pattern and its dissipation remarkably well in this case. This may reflect the relatively high frequency of such clearing events, which can enable data-driven models to learn the associated spatial and temporal evolution from the training data. In contrast, the purely advection-based optical flow model SolarSTEPS-pa performs poorly, as it can only advect existing cloud structures and cannot represent cloud dynamics. When cloud-evolution mechanisms are included, as in SolarSTEPS, the model better captures the observed clearing process. The physics-based IFS-ENS forecast correctly predicts the presence of clouds over the Swiss Plateau. However, its comparatively coarse spatial and temporal resolution limits its ability to resolve cloud edges and the detailed evolution of the dissipation process.

Figure \ref{fig:comparison_inference_power_3} presents the corresponding PV power measurements and forecasts. The measurements, shown in the top row, exhibit spatial and temporal patterns closely aligned with the cloud distribution. In fact, regions affected by clouds correspond to strongly reduced PV power generation, whereas cloud-free areas show enhanced power output. By 10:00 UTC, most PV stations generate power above their respective $P_{95}$ values. Between 08:00 and 09:45 UTC, the national PV power output increased by approximately 72.0\%. Overall, the forecast models reproduce the main SSI patterns, with low-irradiance areas translating into reduced PV power output and vice versa. Consistent with the SSI results, SHADECast and IrradianceNet provide the most accurate power forecasts in this case.

\section{Conclusions}\label{sec:conclusions}
This study presents a comprehensive assessment of solar ramp events and their implications for short-term PV power forecasting. We analyze two years of 15-minute resolution PV power measurements from 6434 PV stations and introduce a quantitative, data-driven methodology to identify solar ramp events, defined as large variations in nationally aggregated PV generation over short time intervals. Within this framework, we systematically characterize the frequency, magnitude, seasonal distribution, and diurnal occurrence of the ramp events. Furthermore, we employ a recently developed spatiotemporal forecasting framework to evaluate both deterministic and probabilistic PV power forecasts under ramp and non-ramp conditions. The benchmarked models, that is SolarSTEPS, SHADECast, IrradianceNet, and IFS-ENS, rely on optical-flow techniques, deep-learning frameworks and physics-based approaches, 

First, our results indicate that solar ramp events occur mainly in spring and summer, with peak frequency between 07:00–09:00 and 14:00–16:00 UTC. In particular, ramp-up events typically result from the combined effect of increasing solar elevation and rapid cloud dissipation during the morning hours, whereas ramp-down events are primarily associated with decreasing solar elevation and enhanced cloud formation in the afternoon. These findings highlight the strong coupling between mesoscale cloud dynamics and variability in nationally aggregated PV power production.

Our study indicates that SolarSTEPS-pa, i.e. the optical-flow-based model relying purely on advection, experiences the largest increase in forecast errors during ramp events, due to its lack of representation of cloud formation and dissipation. Among the satellite-based approaches, SHADECast provides the most reliable forecasts under these conditions. In contrast, under non-ramp conditions, SolarSTEPS performs slightly better than SHADECast for lead times exceeding 60 minutes. Furthermore, the bias-corrected IFS forecasts consistently outperform their raw counterparts, underlining the important role of statistical post-processing.

Quantitatively, satellite-based models exhibit a clear degradation in performance during ramp events compared to non-ramp cases. Specifically, the nRMSE increases by approximately 20–30\% at short lead times, reaching up to 60\% at longer forecast horizons. A similar pattern is observed for nCRPS, with relative increases approaching 70\% beyond lead times of 90 minutes. These findings indicate that ramp events significantly reduce not only deterministic forecast accuracy but also the reliability and sharpness of ensemble predictions. While IFS-ENS also experiences reduced accuracy during ramp events, the magnitude of deterioration is substantially smaller than that of the satellite-based methods. 

The data-driven models benchmarked in our study have been trained on satellite-derived SSI fields in which ramp events are very sparse. Accordingly, the models were not optimised for forecasting rapid SSI changes like the ones examined here. Hence, the models performance in ramp conditions can likely be improved by focusing the model training on ramp events, for instance, by adapting the loss function to place greater emphasis on extreme events. Moreover, future research can involve increasing the fraction of ramp events in the training set, for example, by upsampling and image augmentation of SSI field sequences related to ramp events. Overall, accurately capturing rapid cloud formation and dissipation remains a key challenge for nowcast models. Improving high-resolution spatiotemporal modelling capabilities and increasing the robustness of data-driven approaches to extreme events are essential steps toward enhancing ramp prediction skill. Such advancements are relevant for maintaining grid stability and supporting the reliable integration of growing shares of solar generation into power systems. Future research should also explore alternative spatial aggregation strategies, for example, at the level of transmission grid zones, to better align meteorological variability with operational grid constraints. Moreover, the integration of multi-source observations (e.g., satellite, radar, and ground-based measurements) and the explicit modelling of extreme-event statistics represent promising directions to further improve the prediction of solar ramp events.

\section*{Declaration of Competing Interest}
The authors declare no conflict of interest.

\section*{Data availability}
The authors do not have permission to share the PV data due to a non-disclosure agreement with the data provider.

\section*{Acknowledgements}
We acknowledge funding from the Swiss National Science Foundation (grant 200654). We thank Mathieu Schaer and Christian Steger of MeteoSwiss for the support to account for topographic shading using the HORAYZON library. We also thank Anke Tetzlaff and Uwe Pfeifroth of MeteoSwiss and the German Weather Service, respectively, for compiling and providing the HANNA dataset.

\section*{Declaration of generative AI and AI-assisted technologies in the manuscript preparation process}
During the preparation of this work, ChatGPT was used to help with writing style. After using this tool, the author reviewed and edited the content as needed and takes full responsibility for the content of the published article.

\appendix

\section{Quantitative forecast analysis for the selected case studies}\label{app:appendix_a}
\begin{figure}[t]
	\centering
	\includegraphics[width=1\textwidth]{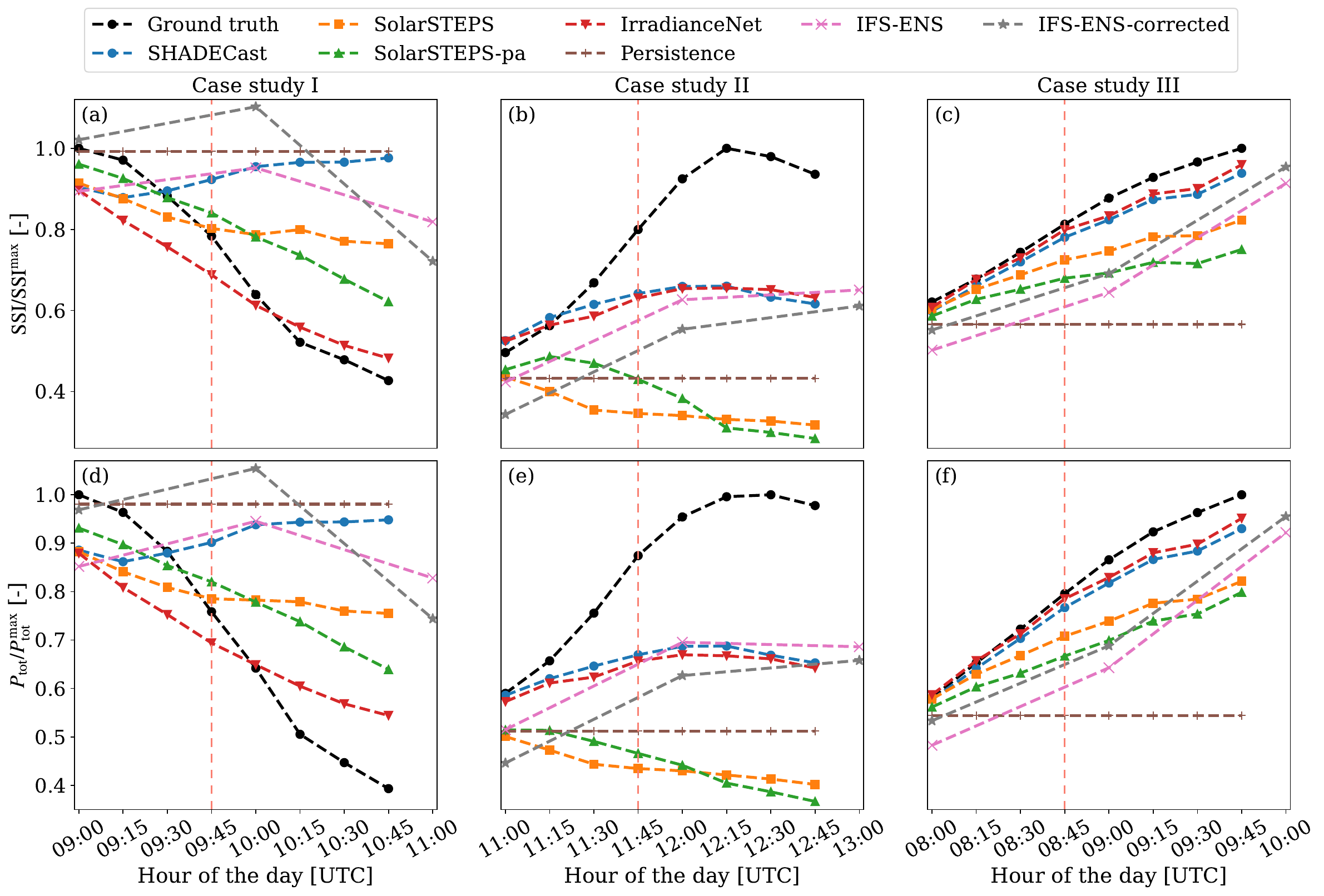}
	\caption{Comparison of (a-c) SSI and (d-f) aggregated PV power forecasts computed for the three case studies discussed in Section 5. The SSI is spatially interpolated to each station location, averaged across all stations, and normalized by its maximum value within the corresponding 2-hour period. The PV power generated is summed over all stations and normalized by the maximum measured aggregated PV power within the same time window. The vertical dashed red line marks the occurrence time of the ramp event.}
	\label{fig:ramp_event_single_forecast_all_comparison}
\end{figure}

This section provides a quantitative comparison of the forecasts for the three case studies introduced in Section \ref{sec:results}, focusing on both SSI and PV power predictions. Figure \ref{fig:ramp_event_single_forecast_all_comparison}(a–c) presents SSI forecasts, interpolated to PV station locations and averaged across ensemble members and stations, against satellite observations, while Figure \ref{fig:ramp_event_single_forecast_all_comparison}(d–f) shows the corresponding aggregated PV power generation. As expected, SSI and PV power exhibit highly correlated temporal behavior, leading to similar trends as a function of the lead time. Case study I is dominated by cloud advection, resulting in a pronounced decrease in SSI and PV production. IrradianceNet provides the most accurate representation of both magnitude and timing. Optical-flow-based methods also capture this decreasing trend reasonably well. In contrast, IFS-ENS significantly underestimates the sharp transition, while SHADECast fails to capture the ramp event, leading to a PV power prediction at two-hour lead time that is more than twice as high as the observed generation. Case study II is characterized by rapid cloud dissipation, resulting in a steep increase in SSI and PV power output. Optical-flow methods struggle in this regime, as they are inherently limited in representing cloud formation and dissipation processes, resulting in persistently underestimated forecasts. Although IFS-ENS, SHADECast, and IrradianceNet capture the overall upward trend, the SSI and PV power variations are notably smoother than the ones observed. For example, at a two-hour lead time, IrradianceNet underestimates PV production by 34.3\%. Finally, case study III involves the dissipation of low-level clouds over the Swiss Plateau, a common meteorological situation in Switzerland. In this scenario, data-driven models perform particularly well, accurately capturing the ramp-up in both SSI and PV power. SHADECast and IrradianceNet show only minor underestimation at a two-hour lead time, with errors of approximately 4.9\%. While IFS-ENS also performs reasonably well, optical-flow methods, despite capturing the increasing trend, tend to underestimate the increase of both SSI and PV power generation.

\section{Metrics}\label{app:appendix_b}
This section introduces the verification metrics used to evaluate the PV power forecasts in Section \ref{sec:results_1}. We consider forecasts for a generic PV installation, which consists of $L=8$ lead times and $E=10$ ensemble members. We denote with $\hat{y}_{s,n,e,l}$ the predicted PV power at station $s$, time step $n$, ensemble member $e$ and lead time $l$. Here, $s \in \{1, \dots, S\}$ indexes the PV installations, $n \in \{1, \dots, N\}$ indexes the time steps, $e \in \{1, \dots, E\}$ indexes the ensemble member and $l \in \{1, \dots, L\}$ indexes the lead time. The corresponding measured PV power at station $s$ and lead time $l$, associated with forecast $n$, is denoted by $y_{s,n,l}$.

Deterministic performance metrics are computed using the ensemble mean forecast, defined as
\begin{equation*}
\overline{\hat{y}}_{s,n,l} = \frac{1}{E} \sum_{e=1}^{E} \hat{y}_{s,n,e,l}.
\end{equation*}
For deterministic models, $E=1$, and thus $\overline{\hat{y}} = \hat{y}$.

The normalized root mean square error (nRMSE) averaged over all lead times and all forecasts for a station $s$, is defined as
\begin{align*}
\text{nRMSE}_s &= \frac{1}{f_{s}} \sqrt{ \frac{1}{N L} \sum_{n=1}^{N} \sum_{l=1}^{L} \big( \overline{\hat{y}}_{s,n,l} - y_{s,n,l} \big)^2 },
\end{align*}
where $f_{s} = P_{95,s}$. This normalization ensures comparability across stations, as installations with larger $P_{95,s}$ would otherwise exhibit systematically higher errors.

In addition to deterministic metrics, we evaluate the normalized continuous ranked probability score (nCRPS), which assesses the reliability and sharpness of the full predictive distribution. The nCRPS compares the predictive cumulative distribution function $F_{s,n,l}$, derived from the ensemble forecast, with the observation $y_{s,n,l}$, and is defined as
\begin{equation*}
\text{nCRPS}_s = \frac{1}{N L f_s} \sum_{n=1}^{N} \sum_{l=1}^{L} 
\int_{-\infty}^{\infty} \big( F_{s,n,l}(z) - \mathds{1}(y_{s,n,l} \leq z) \big)^2 \, dz,
\end{equation*}
where $\mathds{1}(\cdot)$ denotes the indicator function. In practice, $F_{s,n,l}$ is approximated from the empirical distribution of the ensemble members $\{\hat{y}_{s,n,e,l}\}_{e=1}^E$.

Both deterministic and probabilistic metrics can be further averaged across all stations, yielding the global nRMSE and nCRPS illustrated in Figure~\ref{fig:power_comparison_line_ramp_yes}. 

\clearpage

\bibliographystyle{elsarticle-num} 
\bibliography{BibTex}

\end{document}